\documentclass[runningheads, envcountsame, a4paper]{llncs}

\usepackage[numbers]{natbib}

\usepackage[utf8]{inputenc} 
\usepackage[T1]{fontenc}    
\usepackage[hyphens]{url}
\usepackage{booktabs}       
\usepackage{amsfonts}       
\usepackage{nicefrac}       
\usepackage{microtype}      
\usepackage{tikz}
\usepackage{amssymb,amsmath,amsbsy}
\usepackage{multicol}
\usepackage{verbatim}
\usepackage{cleveref}
\usepackage{bm}
\usepackage{wrapfig}
\usepackage{microtype}

\usepackage[boxed]{algorithm2e}

\newcommand{\x}{\mathbf{x}}

\newcommand{\y}{\mathbf{y}}

\newcommand{\lik}{\mathcal{L}}
\newcommand{\nlaw}{\mathcal{N}}

\newcommand{\Rset}{\mathbb{R}}
\newcommand{\Xset}{\mathbb{X}}
\newcommand{\Nset}{\mathbb{N}}

\newcommand{\new}{_{\text{new}}}

\newcommand{\esp}{\mathbb{E}}

\DeclareMathOperator*{\var}{Var}

\newcommand{\truvar}{\textsc{Tru}\textsc{Var}}

\let\originalleft\left
\let\originalright\right
\renewcommand{\left}{\mathopen{}\mathclose\bgroup\originalleft}
\renewcommand{\right}{\aftergroup\egroup\originalright}

\newcommand{\GP}{\ensuremath{\mathcal{GP}}}

\DeclareMathOperator*{\argmax}{\operatorname{arg\,max}}
\DeclareMathOperator*{\given}{|}
\DeclareMathOperator{\E}{\mathbb{E}} 
\renewcommand{\Re}{\mathbb{R}} 
\newcommand{\KL}{\operatorname{KL}}
\DeclareMathOperator{\Gauss}{\mathcal{N}} 

\renewcommand{\vec}[1]{\bm{\mathrm{#1}}} 
\newcommand{\mat}[1]{{\mathrm{#1}}} 

\DeclareMathOperator*{\argmin}{arg\,min}

\usepackage{xspace}

\newcommand{\MX}{\mat{X}}

\newcommand{\MZ}{\mat{Z}}

\newcommand{\MS}{\mat{S}}

\newcommand{\MK}{\mat{K}}

\newcommand{\vu}{{\vec{u}}}

\newcommand{\vm}{{\vec{m}}}
\newcommand{\vz}{{\vec{z}}}
\newcommand{\vk}{{\vec{k}}}

\begin{document}

\title{Automatic Tuning of Stochastic Gradient Descent with Bayesian Optimisation}

\titlerunning{Automatic Tuning of SGD with BO}

\author{Victor Picheny \and Vincent Dutordoir \and Artem Artemev \and Nicolas Durrande}

\authorrunning{V. Picheny et al.}

\institute{PROWLER.io, 72 Hills Road, Cambridge, CB2 1LA, UK\\ \email{$\{$victor,vincent,artem,nicolas$\}$@prowler.io}}

\maketitle    

\begin{abstract}
Many machine learning models require a training procedure based on running stochastic gradient descent. A key element for the efficiency of those algorithms is the choice of the learning rate schedule. 
While finding good learning rates schedules using Bayesian optimisation has been tackled by several authors, adapting it dynamically in a data-driven way is an open question.
This is of high practical importance to users that need to train a single, expensive model.
To tackle this problem, we introduce an original probabilistic model for traces of optimisers, based on latent Gaussian processes and an auto-/regressive formulation,
that flexibly adjusts to abrupt changes of behaviours induced by new learning rate values. 
As illustrated, this model is well-suited to tackle a set of problems:
first, for the on-line adaptation of the learning rate for a cold-started run; 
then, for tuning the schedule for a set of similar tasks (in a classical BO setup), as well as warm-starting it for a new task.

\end{abstract}

\section{Introduction}

The great recent successes of machine learning 
generally rely on models with high complexity (e.g.~deep models) and extensive datasets.
Those models usually require running a training procedure over a (large) set of parameters,
which often amounts to minimising a loss function with an iterative algorithm such as stochastic gradient descent (SGD) or 
the non-linear conjugate gradient method.
In the deep learning community, a particular focus has been given to SGD algorithms such as
Adagrad \citep{duchi2011adaptive}, RMSProp \citep{tieleman2012lecture}, and in particular Adam \citep{andrychowicz2016learning}, 
which despite recent discussions and improvements \citep{loshchilov2018decoupled,reddi2018convergence} can be considered as the state-of-the-art and is widely used in practice \citep{gugger2018}.
Unfortunately, this training procedure is often extremely time-consuming due to the high model complexity and the amount of data at hand;  
hence, performing it with the best possible efficiency is paramount for many applications,
in particular those for which training is done repeatedly, or continuously, for example in streaming models.

The performance of nearly all SGD variants depend critically on the choice of the \emph{learning rate} level \citep{bengio2012practical},
which in short tunes by how much the algorithm should follow the (noisy) gradient signal.
The default choice for most algorithms is a constant learning rate, 
although recent experiments showed that a time-varying value can be extremely beneficial 
\citep{baydin2017online,smith2017cyclical,smith2019super}.
In any case, learning rate values require to be set, 
and tuning them by hand can be excessively burdensome.
Bayesian optimisation (BO), on the other hand, is now a well-established tool for model selection and parameter tuning \citep{bergstra2011algorithms,snoek2012practical}.
While learning rates are sometimes included in the parameters to be tuned \citep{bergstra2011algorithms},
to our knowledge no work has been dedicated to speeding up SGD algorithms on-the-fly using BO: this is the purpose of the present work.

Let $\lik: {\Theta} \times {\Omega} \rightarrow \Rset$ be the objective optimised by SGD; 
typically in deep learning, $\lik$ can be a mean squared error or an evidence lower bound (ELBO)\footnote{In the following, without loss of generality we use the convention that $\lik$ should be maximised.}.
$\theta \in {\Theta}$ denotes a set of parameters, while $\omega \in \Omega$ defines the learning task at hand,
that is, a particular model structure and a dataset. 
$\Omega$ is typically a singleton (when a single model is fitted to a single dataset),
or a discrete set, for instance when the same model is used to fit different datasets.
Given $\Omega$, an SGD algorithm produces a sequence of parameters $\theta_0, \ldots, \theta_T$,
with $T \in \Nset^*$ a pre-defined number of iterations (i.e. optimisation steps).

In our setup we assume that the learning rate varies from one iteration to the next.
One way to parametrise it, which avoids working directly on a high-dimensional vector of size $T$, is to use a piecewise constant form. 
Defining a sequence of length $d+1 \ll T$:
\begin{equation*}
 T_0 =0 < T_1 < \ldots < T_d = T,
\end{equation*}
the learning rate curve $\gamma(t)$ is defined by $d$ constants $x_0, \ldots, x_{d-1}$ such that
\begin{eqnarray*}
 \gamma(t) = x_k \text{ for } T_{k} \leq t < T_{k+1}, \quad \forall t \in [0, T], \quad
 \forall 0 \leq k < d.
\end{eqnarray*}
Assuming lower and upper bounds for $x_k$, without loss of generality our tuning parameters ${\x \in \Xset}$ can be rescaled to $[0, 1]^d$.
For any task $\omega \in \Omega$, our objective is to seek the best possible learning rate, that is, the one that maximises $\lik$ after $T$ steps:
\begin{equation}
 \x_{\omega}^* = \argmax_{\x \in \Xset}  \lik\left[\theta_T\left(\x, \omega\right), \omega \right],
\end{equation}
with $\theta_T(\x, \omega)$ the parameters returned after $T$ iterations (now written as a function of the learning rate schedule $\x$ and task $\omega$).

Now, BO classically relies on Gaussian process (GP) surrogate models of the function to optimise. 
While predicting trace (that is, the value of $\lik$ for all $t$) has been already addressed in related settings \citep{picheny2013nonstationary,domhan2015speeding,klein2016learning},
adapting those models to make them dependent on a learning rate that changes over time
is often not possible. Moreover, changing the learning rate often induces drastic changes on the trace behaviour, which makes typical parametric models unfit for the task.
Our first contribution is the design of an auto-regressive model for the trace  of a SGD algorithm, that flexibly adjusts to abrupt changes of behaviours.
The model, which serves as a base for our sampling strategies, is presented in \cref{sec:model}.

In the classical BO setting \citep{shahriari2016taking}, first an initial set of experiments, typically with space-filling properties, $\{\x_1, \omega_1\}, \ldots, \{\x_{N_0}, \omega_{N_0}\}$ is run. The gathered data would then be used to build a predictive model for $\lik\left[\theta_T\left(\x, \omega \right), \omega\right]$, which would allow us to estimate $\x_{\omega}^*$. In BO,
such estimate is gradually improved by designing a sequence of additional experiments $\x_{N_0+1}, \ldots, \x_N$ that enhances the prediction capability of the model, following an exploration / exploitation trade-off.

However, often times the user is faced with training a single model, without much prior information on an appropriate choice of the learning rate schedule. In that case, the relevance of BO is somehow debatable, as it is uncertain that the outcome of $N$ SGD runs would outperform a single but much longer one with an empirically chosen learning rate. 
To overcome this,
\citep{swersky2014freeze,li2018hyperband,falkner2018bohb} used forecasting models to stop early unpromising runs and drastically reduce the computational cost of the BO loop. Yet, such approaches can only provide long-term recommendation at the price of repeated long runs, and are mostly relevant when other parameters are tuned simultaneously with the learning rate. 

We propose here an alternative strategy, unexplored in the BO literature,
which is to adapt a single run \textit{on the fly} rather than terminate some and start new ones from scratch.
In addition, as modern hardware architecture favours parallel computing, 
we want to leverage the fact that training procedures can be run in parallel (say, one on each available GPU).
\Cref{sec:coldstart} is dedicated to this problem.

In other practical situations, training is done repeatedly
over similar tasks: for instance the same model fitted to different datasets or different variations of a model fitted to a dataset. In that case, one seeks an optimal mapping \citep[or profile optima,][]{ginsbourger2014bayesian} $\Phi: \Omega \rightarrow \Xset$
(from the space of tasks to the space of learning rates), such that 
$\forall \omega \in \Omega, \Phi(\omega) = \x^*_{\omega}$.
Transferring information from one task to another allows designing strategies much more efficient than running 
independent BO loops for each task \citep{swersky2013multi}. 
This framework is considered in \cref{sec:multitask}.

\section{A GP-based NARX Model For Optimisation Traces}\label{sec:model}
For simplicity, we focus for now on the case where $\Omega$ is a singleton (i.e. we consider a single dataset and model) and remove the dependence on $\omega$ from our notations.
The generalisation to multiple tasks is deferred to \cref{sec:multitask}. 

\subsection{Modelling of Optimisation Traces}
Define first
\begin{eqnarray*}
 y\left(\x, t \right) =\lik\left[\theta\left(\x, t\right) \right],
\end{eqnarray*}
the trace of an optimisation run and denote by $Y_k = y(\x, T_k)$ the trace value at each changing point for the learning rate.

\begin{figure*}[ht]
 \centering
  \includegraphics[width=.32\textwidth]{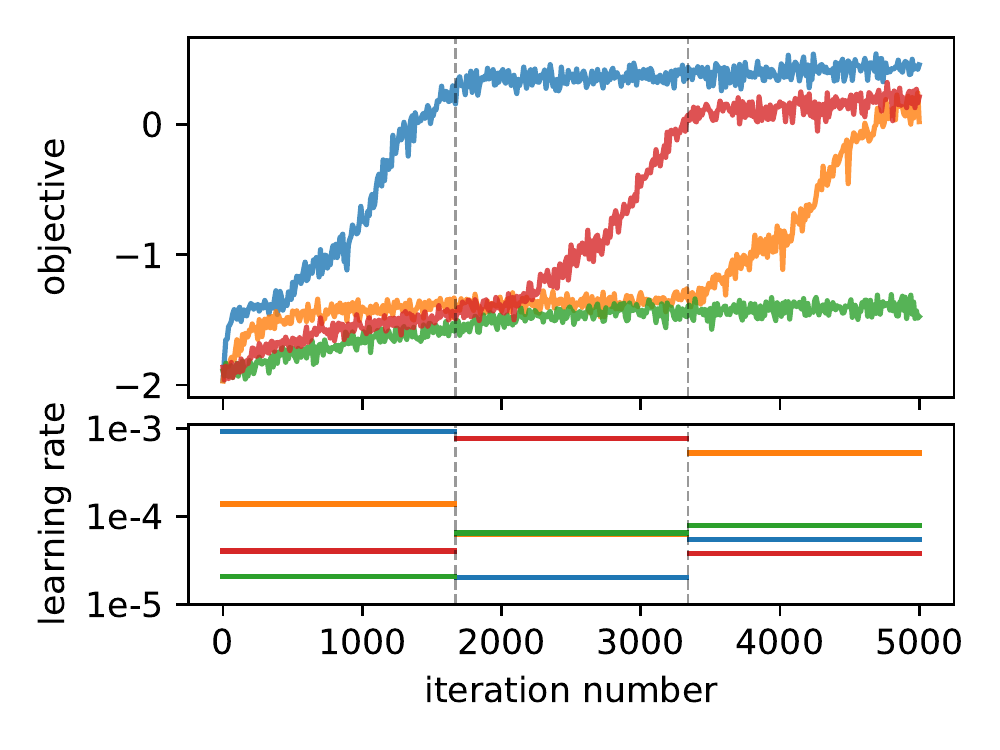}
  \includegraphics[width=.32\textwidth]{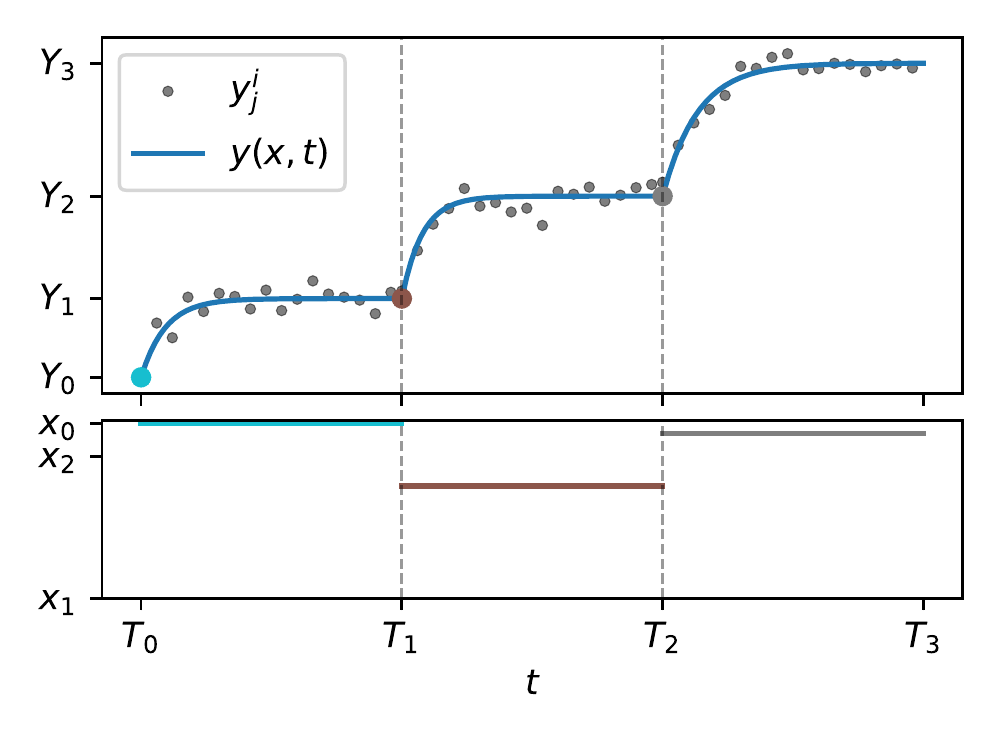}
  \includegraphics[width=.32\textwidth]{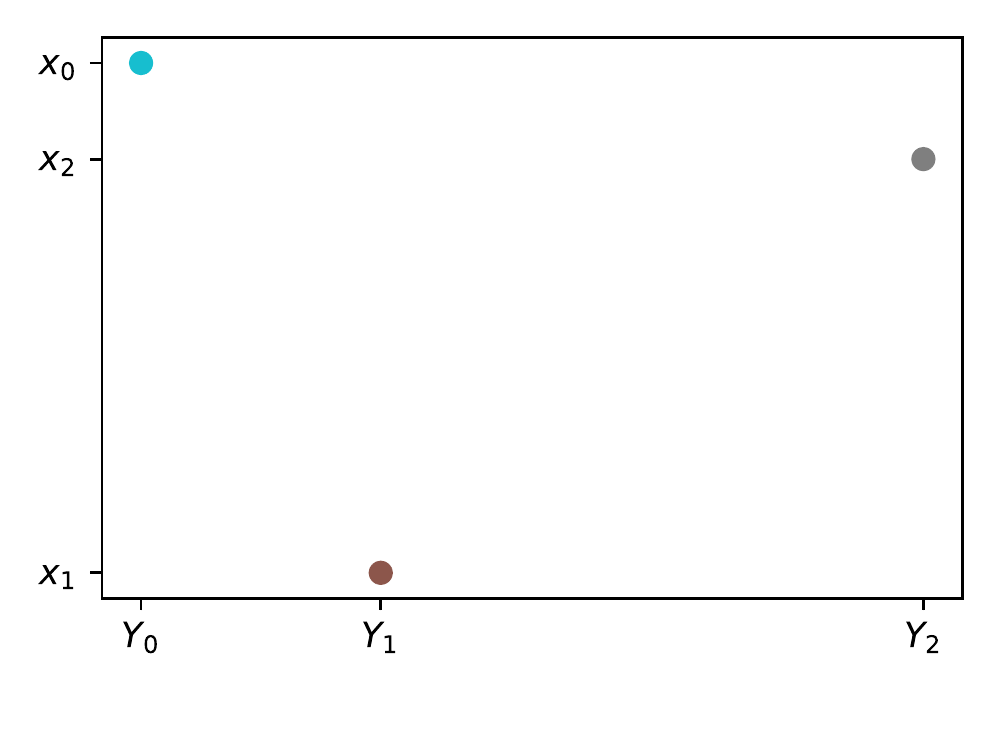}
  \caption{Left: four traces (top) corresponding to the training of an SVGP model using Adam and corresponding learning rates. Middle: notations and model. Right: corresponding data point locations for the latent GP.} \label{fig:formatting}
\end{figure*}

There exist many options to fit a parametric model to $y\left(\x, t \right)$: for instance,
in \cite{domhan2015speeding}, 11 models for traces are proposed, including exponential forms 
($c - \exp(-a t^\alpha + b)$) and power ones ($c - a t^\alpha$). 
While those forms make sense with a constant
learning rate, they cannot fit properly a varying one, as changing the learning rate drastically modifies the trace dynamics. 
This is illustrated in \cref{fig:formatting}, left, where the  visible trace trend abruptly changes every time the learning rate changes 
(see also \cref{fig:multiresults}).
Hence, instead of fitting a single parametric model, we propose to use a composite one, based on an auto-regressive formulation.

First, we model $Y_0$ as an i.i.d. Gaussian variable $\nlaw(m_0, \sigma_0^2)$, 
to account for randomness in the starting point.
Then, we propose to model the trace using a
non-linear auto-regressive model with exogenous inputs \citep[NARX(q),][]{leontaritis1985input}, 
which general expression is:
\begin{equation*}\label{eq:narx}
 Y_{k+1} = \Gamma \left(\{Y_{k - l}\}_{l=0}^{q-1}, \{x_{k - l}\}_{l=0}^{q-1} \right) + \varepsilon_k, 
\end{equation*}
where the new state is a non-linear function of the $q$ past states
and exogenous inputs plus some independent noise.
We propose a specific function for $\Gamma$ as follow: 
\begin{equation*}\label{eq:recursive_def}
    Y_{k+1} = Y_k + \eta \left[ f \left(\{Y_{k - l}\}_{l=0}^{q-1}, \{x_{k - l}\}_{l=0}^{q-1} \right), T_{k+1} - T_k \right] + \varepsilon_k,
\end{equation*}
where $f: \Rset^{2q} \rightarrow \Rset^p$ is a latent function that modulates the increments according to the current and past trace values $\{Y_{k - l}\}_{l=0}^{q-1}$ and learning rates $\{x_{k - l}\}_{l=0}^{q-1}$ and
$\eta: \Rset^{p+1} \rightarrow \Rset$
(or $\Rset^+$ to ensure the monotonicity of the trace) is a link function that returns the trace increment for any given time $t$ according to its parameters.
As the trace may be recorded for $t$ values outside ${T_0, T_1, \ldots}$, our model for the trace is:
\begin{equation*}
  y(\x, t) = Y_k + \eta \left[ f\left(\{Y_{k - l}\}_{l=0}^{q-1}, \{x_{k - l}\}_{l=0}^{q-1} \right), t - T_k \right] + \varepsilon(t),
\end{equation*}
where $T_k \leq t < T_{k+1}$ and $\varepsilon(t)$ represents the noise in the case where the objective function is not evaluated exactly and/or for unaccounted-for deviations between observations and the model.

In the following, we use either piecewise linear or piecewise exponential forms for the traces. With respectively $f=f_1: \Rset^{2q} \rightarrow \Rset$ defining the linear slope and $f=(f_1, f_2): \Rset^{2q} \rightarrow \Rset^2$ with $f_1$ corresponding to a logit offset and $f_2$ to a logit rate, this gives:
\begin{align}
    \eta_{\text{lin}}(f, t) &= \phi \left(f_1 \right)\,t, \label{eq:lin-link}\\
  \eta_{\exp}(f, t)  &= \phi\left(f_1 \right)\,\left(1 - \exp \left[-\phi \left( f_2 \right)\,t \right] \right),\label{eq:exp-link}
\end{align}
where $f$ is a (multi-output) GP and $\phi(\cdot)$ is here the softplus function, $\phi(u) = \log(1 + e^u)$, to ensure monotonicity.
Note that any of the parametric models of e.g. \cite{domhan2015speeding} may be used here as $\eta$. In a sense, our model extend those for a time-varying learning rate.

\begin{figure}[htbp]
\includegraphics[trim= 15mm 10mm 10mm 15mm, clip, width=.48\textwidth]{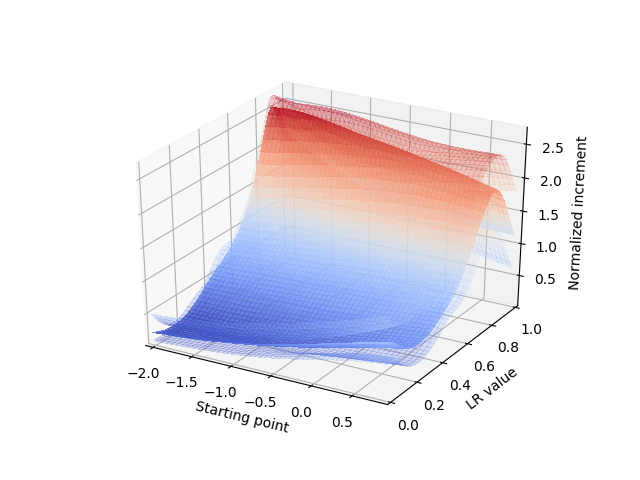}
\includegraphics[trim= 15mm 10mm 10mm 15mm, clip, width=.48\textwidth]{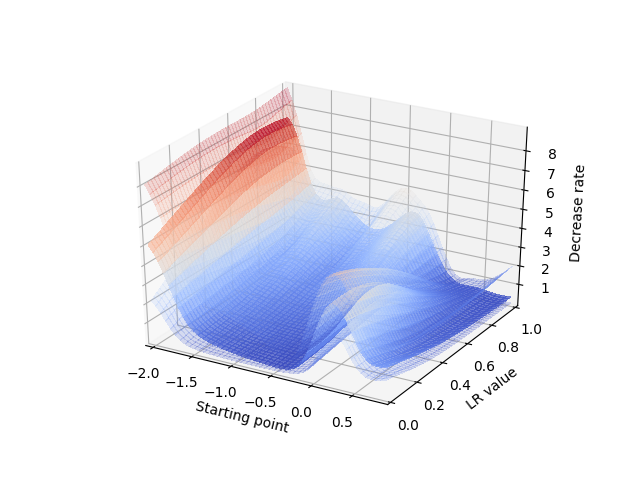}
 \caption{Consider the NARX(1) model of \cref{sec:model} with the exponential link function of \cref{eq:exp-link}. On the left, we plot the mean $\pm$ the empirical std. dev. of the GP for the increment factor $\phi(f_1)$. On the right, we plot the GP surface of $\phi(f_2)$ which models the decay rate of the exponential. Given $q=1$, both GPs are only a function of the most previous trace value `starting point' and the current learning rate `LR value'. These surfaces are learned as part of experiment \cref{sec:exp:mnist}. We notice, for example, that for low starting values of the objective, the response $\eta(f, t)$ behaves like a step-function.} \label{fig:model}
\end{figure}

Focusing on the linear case, and setting $q=1$, we have: 
$y(\x, t) = Y_k + \phi \left[f_1 \left(Y_k, x_k \right) \right] \left(t - T_k \right)$.
We directly see the dynamic implied by our model: increments are linear with respect to time, but
the slope depends non-linearly on an exogenous input (the learning rate) and on the current state (intuitively, we expect flatter slopes if $Y_k$ is high, as the model is close to convergence). Considering the exponential case (\cref{eq:exp-link}) and setting $q = 1$, we show GP surfaces in  \cref{fig:model} for $\phi(f_1)$ (left) and $\phi(f_2)$ (right).

Importantly, the same $\eta$ is used over all the time intervals and only depends on $t - T_k$.
Hence, in the NARX(1) case, our model implies that the same initial conditions and learning rates would lead to the same outcome, regardless of the time the algorithm has been run before.
In that sense, the model can be considered as Markovian, 
as it is independent of the path followed by the optimiser before $T_k$.
While our model is general for higher-order Markov chains (i.e. NARX(q)), 
we found empirically that the current formulation provides the right trade-off between accuracy, robustness and ease of inference, as the link function $\eta$ is only defined over $\Rset^{2}$.

Finally, we follow the \textit{chained GP} framework \citep{saul2016chained} for $f$, and choose a GP prior with independent components.
\subsection{Learning $f$ Using Variational Inference}\label{sec:inference}
We consider here a fixed link function, so the inference task boils down to estimating the parameters of the function $\eta$ and $Y_0$ (the starting value of the trace).
Assume that we have run the optimiser $N \geq 1$ times for different learning rate schedules $\x^1, \ldots, \x^N$ 
and recorded the trace
of each run $i$ at times $t_1^i, \ldots, t_{m_i}^i$ (in $[T_0, T_d]$).
Let
\begin{equation*}
    y_j^i = y(\x^i, t_j^i) \quad (1 \leq i \leq N, 1 \leq j \leq m_i)
\end{equation*}
denote an observation of the trace at time $t_j^i$ for a learning rate schedule $\x^i$. 
We assume that each trace is evaluated at the beginning of each time interval 
(i.e. $\forall i, k , \exists j \text{ s.t. } t_j^i = T_k$),
which gives explicit observations of $Y_k$, which we denote henceforth $Y_k^i$.

The parameters of $Y_0 \sim \nlaw(m_0, \sigma_0^2)$ can be inferred by maximum likelihood from $\{Y_0^1, \ldots, Y_0^N\}$.
Now, we focus on learning the latent GP functions $f_1, f_2, \ldots f_p$. 
Without loss of generality, we assume in the following that our link function only requires a single GP parameter $f$ (i.e. p = 1). When a link function depends on more than one GP we simply apply the same procedure in parallel and model every function independently.

\newcommand{\tX}{\tilde{\MX}}
\newcommand{\tx}{\tilde{\x}}
\newcommand{\Ktxtx}{\MK_{\tX\tX}}

We follow a fully Bayesian approach to infer $f$ from the given dataset. 
We place a GP prior on the latent function $f$ and assume that $\varepsilon$ is i.i.d. Gaussian noise with zero mean and $\sigma^2$ variance, so that:
\begin{eqnarray*}
    f &\sim& \GP\big(0, k(\cdot, \cdot)\big) \quad \text{and} \\
    y^i_j \given f, Y_{k}^i, x^i_k, t^i_j
    &\sim& \nlaw\left(y_j^i \given \eta(f(x^i_k, Y_{k}^i), t^i_j - T^i_k), \sigma^2\right),
\end{eqnarray*}
with $k$ such that $T_{k} \le t^i_j < T_{k+1}$.
For general link functions $\eta$ is exact inference of the latent function $f$ not possible due to the non-linear transformation.

A classical solution is to follow the Sparse Variational GP (SVGP) framework \citep{titsias2009, hensman2015scalable}, that relies on two components.
First, it introduces a set of \emph{inducing variables}, which main use is to specify the function value of the posterior GP at a specific set of $m$ pseudo inputs $\MZ = \{\vz_i\}_{i=1}^m$, denoted as $\vu = f(\MZ)$. 
The distribution of the inducing variables is specified by a fully parameterised Gaussian $q(\vu) = \Gauss(\vm, \MS)$
with mean $\vm \in \Re^m$ and covariance $\MS \in \Re^{m\times m}$, which are the variational parameters we want to learn. The prior GP on $f$ can then be conditioned on $\vu$, which leads to a marginal posterior $q(f)$ with mean $\mu(\cdot)$ and the variance $\Sigma(\cdot, \cdot)$:
\begin{align}
\label{eq:qf}
   \mu(\cdot) &= \vk_{\MZ}^\top(\cdot) \MK_{\MZ\MZ}^{-1} \vm \quad\text{and}\nonumber \\
   \Sigma(\cdot, \cdot) &= k(\cdot, \cdot) + \vk_{\MZ}^\top(\cdot) \MK_{\MZ\MZ}^{-1}(\MS - \MK_{\MZ\MZ})\MK_{\MZ\MZ}^{-1} \vk_{\MZ}(\cdot),
\end{align}
where $\vk_{\MZ}(\cdot) := \left[k(\vz_i, \cdot)\right]_{i=1}^m \in \Re^m$ and $[\MK_{\MZ\MZ}]_{ij} = k(\vz_i, \vz_j)$. 

With this approximation in place we can set up our model's optimisation objective, which is a lower bound on the log marginal likelihood \citep[ELBO,][]{hoffman2013}, equal to
\begin{equation}
        \sum_{i=1}^N \sum_{j=1}^{m_i} \E_{q(f)}\big[ \log \nlaw\left(y_j^i \given \eta(f(x^i_k, Y_{k}^i), t^i_j - T^i_k), \sigma^2\right) \big] - \KL\big[q(\vu) || p(\vu)\big], \label{eq:elbo}
\end{equation}
where $k\text{ s.t. }T_{k} \le t^i_j < T_{k+1}$, and $\KL$ is the Kullback-Leibler divergence between the approximate and the prior of $f$ \citep{matthews2016sparse}.
It can be calculated analytically given the Gaussianity of both the prior and posterior on $\vu$. 
The expectation can be estimated in an unbiased way using Monte-Carlo, by sampling $q(f)$ (\cref{eq:qf}) and propagating the samples through the link-function.
Optimising \cref{eq:elbo} with respect to the model parameters can be done by gradient descent thanks to automatic differentiation toolkits. 

In our experiments we made use of the GPflow library \citep{matthews2017gpflow}. 
More precisely, we used the provided multi-output framework for GPs \citep{GPflow2020multioutput}, which is well-suited for implementing and optimising of these complex, composite GP models. 

Note that as the approximation is sparse (i.e. it relies on a few inducing points), it can handle much larger datasets than classical GP models. This is a decisive advantage here as traces typically contain thousands of datapoints. 

\subsection{Generating Trace Predictions}\label{sec:prediction}
Since the main objective is to maximise $\lik$ after $T$ iterations, 
we would like to predict 
\begin{equation*}
    Y_d = y\left(\x, T\right) = \eta\left( f(Y_{d-1}, x_{d-1}), T_d - T_{d-1} \right),
\end{equation*}
which requires access to $Y_{d-1}$.
Recursively, we see that predicting $y\left(\x, T\right)$
is achieved by predicting the corresponding sequence $\{Y_0, \ldots, Y_{d} \}$.
Importantly, the distribution of $\{Y_0, \ldots, Y_{d} \}$ is not available analytically
because of the arbitrary link function. Hence, we must resort to sampling.
Given the recursive structure (the value of $Y_i$ is necessary to draw from $Y_{i+1}$) we sample first $Y_0$, then $Y_1$ after conditioning $f$ on $f(Y_0, x_0)$, and recursively sampling 
$Y_{i+1}$ ($1\leq i < d$) after conditioning $f$ on 
$f(Y_0, x_0), \ldots, f(Y_{i}, x_{i})$.
Drawing multiple samples for a given $\x$ may be used to provide any statistic of $y(\x, t)$, such as mean, variance and quantiles. 

\section{Dynamic Tuning of the Learning Rate}\label{sec:coldstart}
We address now the question of dynamically tuning the learning rate
for a single model and task, using the model previously defined.
To do so, we depart from the standard BO approach, 
by modifying a small set of runs on the fly instead of starting repeatedly new ones.

\subsection{Proposed Strategy}
We consider the following framework.
We assume that $Q$ SGD runs are performed in parallel with different learning rates. 
For simplicity, we assume synchronicity (all runs progress with the same speed).
Each run is conveyed independently over time intervals $[T_k, T_{k+1}]$. 
At $t=T_{k+1}$, the $Q$ traces are fed to the model, 
which is then used to schedule $Q$ new learning rates for the next interval.

The learning rates are chosen as follows.
At initialisation, as there is no model to help making decisions, $x_0^1, \ldots, x_0^Q$ are taken on a uniform grid between bounds.
At any $T_k$ ($k>0$), the current trace observations are first integrated into the model. 
Then, each new learning rate $x_{k}^i$ is chosen to maximise the $\alpha_i$-quantile $q_{\alpha_i}$ (computed here empirically by sampling) of the trace at the end of the next interval:
\begin{equation}\label{eq:ucbargmaxpart}
    x_{k}^i = \argmax_{x \in [0, 1]} q_{\alpha_i} \left[ \eta \left( f \left(y_i^{j}, x \right), T_k - T_{k-1}\right) \right].
\end{equation}
The $\alpha_i$'s are used here to balance exploitation and exploration: $\alpha_i$'s close to one may lead to very optimistic choices (learning rates for which the outcome is highly uncertain) while $\alpha_i$'s close to zero result in risk-averse choices (guaranteed immediate performance).
Hence, to maximise our diversity of choices we set $\alpha_1=\tfrac{1}{2Q}, \alpha_2=\tfrac{3}{2Q}, \ldots, \alpha_Q=1 - \tfrac{1}{2Q}$. 
In the case $Q=1$, this forces to choose $\alpha=0.5$, which is a risk-neutral strategy. 
Following an optimistic strategy (say, $\alpha=0.75$) instead may enhance exploration and
improve long-term performance. 

In addition here, at each $T_i$ we greedily select the run with highest current trace and duplicate it $Q$ times while discarding the others. While this was found to accelerate significantly the performance, keeping each run may prove a valid alternative on problems that require more learning rate scheduling exploration.

The pseudo-code of the strategy is given in alg. \ref{alg:coldstart}. 
Note that a relevant choice of $d$ is problem-dependent: a large $d$ allows more changes of the
learning rate value, but increases the computational overhead due to model fitting and solving \cref{eq:ucbargmaxpart}. 
Besides, to facilitate inference the trace may not be sliced in too many parts (\cref{fig:formatting}). In the experiment reported below, using $d=20$ resulted with a negligible BO overhead.

\begin{algorithm}[H]
    \caption{Single task tuning}\label{alg:coldstart}
	Choose $Q$, $d$, $T_d$, set $k=0$\;
    Take $x_0^1, \ldots, x_0^Q$ on a regular grid\;
    Run $Q$ SGDs for $T_1$ steps\; 
    Gather $Q$ traces, read $Y_1^i$'s, and build the model\;
    \For{$k \gets 1$ to $d-1$}{
      Get $i^* = \argmax_{1 \leq i \leq Q} Y_k^i$\;
      Duplicate $i^*$'s SGD run $Q$ times\; 
      $\forall 1 \leq i \leq Q$, find $x_{k}^i$ by solving \cref{eq:ucbargmaxpart}\;
      Pursue duplicated run for $T_{k+1} - T_k$ iterations and learning rates $x_k^1, \ldots, x_k^Q$, resp.\;
      Gather $Q$ traces, collect $Y_k^i$'s, and update model\;
    }
    Find $\argmax_{1 \leq i \leq Q} Y_d^i$, return corresponding parameters\;
\end{algorithm}

\subsection{Experiment: Dynamic Tuning of Learnig Rate on CIFAR}
We apply our approach to the training of a vanilla ResNet \citep{he2016deep} neural network with 56 layers on the classification dataset CIFAR-10 \citep{krizhevsky2009learning} that contains 60,000 32 $\times$ 32 colour images. We use an implementation of the ResNet model for the CIFAR dataset available in Keras \citep{resnetkeras}. We first split the dataset into 50,000 training and 10,000 testing images. The Adam optimiser is used for 100 epochs to maximise the log cross-entropy for future predictions.

\begin{figure*}[tbh]
\centering
\includegraphics[width=\linewidth]{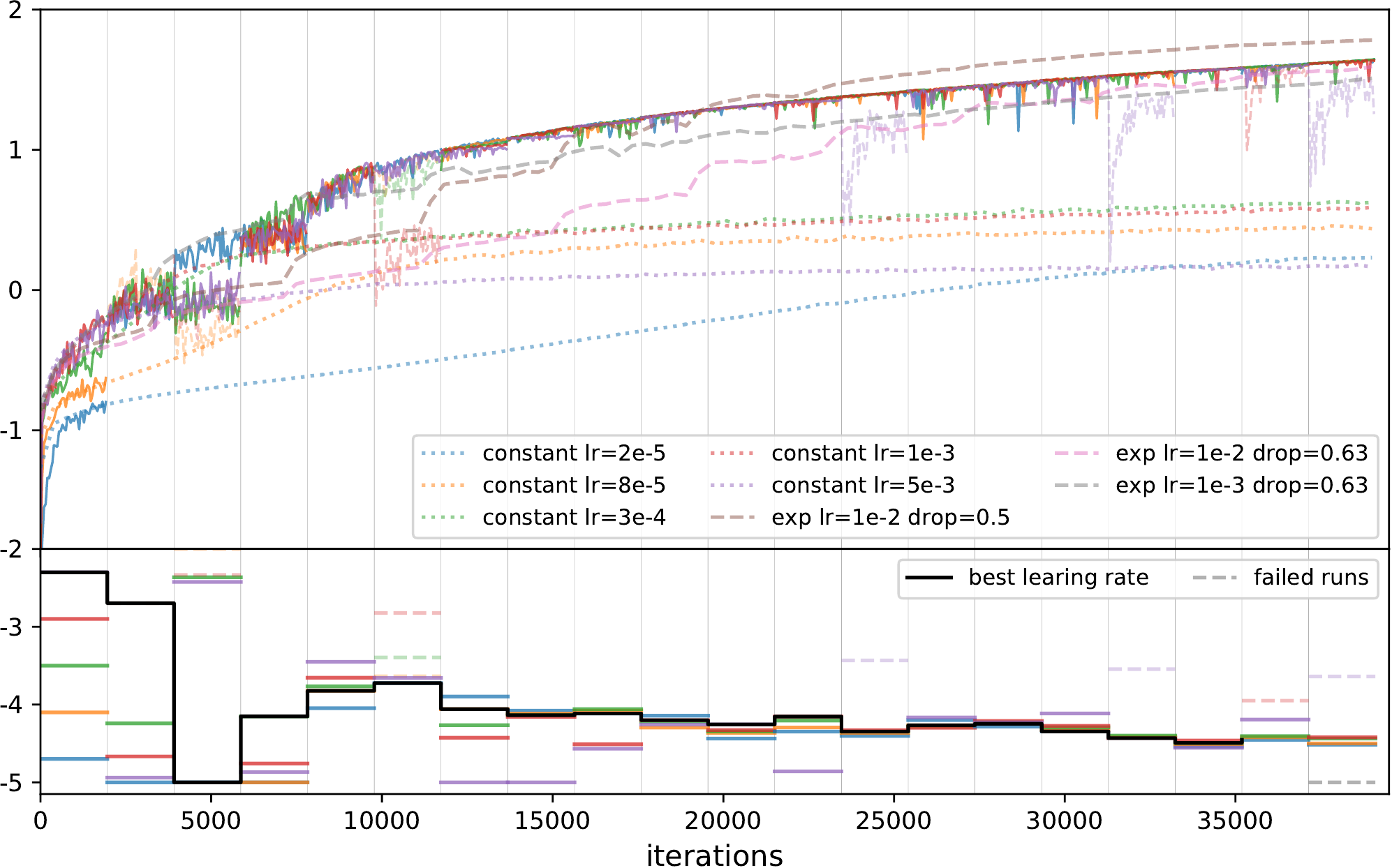}
\caption{Learning rates (bottom) and corresponding traces (top), following Algorithm~1. Dashed lines highlight `failed' runs according to our classifier. The baselines (constant learning rates) are shown by dotted lines and the optimal learning rate schedule is given in black.}
\label{fig:resnet-learning-rates}
\end{figure*}

Our BO setup is as follows: the 100 epochs are divided into $d=20$ equal intervals and $Q=5$ optimisations are ran in parallel. The objective is recorded every 50 iterations.
The GP model for $f$ uses a Mat\'ern-$5/2$ kernel, a linear link function and $100$ inducing points.
A practical issue we face is that increasing abruptly the learning rate sometimes causes aberrant behaviour (which can be seen by large peaks in the trace in \cref{fig:resnet-learning-rates}). 
To avoid this problem, we use a GP classification model \citep{hensman2015scalable} 
to predict which runs are likely to fail based on $\{Y, x\}$ values. The optimisation of \cref{eq:ucbargmaxpart} is then restricted to the values of $x$ for which the probability of failure is lower than $\alpha_j$. We set the threshold for failure for a given trace inverse proportional to its quantile $\alpha_j$ as we want traces with larger $\alpha_j$ be more explorative. In addition, we limit the maximum change to one order of magnitude.

As baselines, we use five constant learning rates schedules uniformly spread in log space between $10^{-5}$ and $10^{-2}$, and 12 learning rates with exponential decay (three initial values $\gamma_0$ between $10^{-4}$ and $10^{-2}$ and four decay rates $\gamma$, $0.5$, $0.63$, $0.77$ and $0.9$), such that the learning rate in each epoch equals $\gamma_0 \times \gamma^{-\text{epoch}/10}$.

\Cref{fig:resnet-learning-rates} shows the dynamic of our approach. The initial interval shows the large performance differences when using different learning rates. Here, a very large learning rate is best at first, but almost immediately becomes sub-optimal. After a quarter of the optimisation budget, the optimal learning rates always takes values around $10^{-4}$, slowly decreasing over time.
The algorithm behaviour captures this, by being very exploratory at first and much less towards the last intervals. 

Comparing to constant learning rate schedules, our approach largely outperforms any of them. In the case where no parallel computation is performed, our approach would still outperform any constant learning rate, as those seem to have converged already to sub-optimal values after 40,000 iterations. Our approach also outperforms all exponential decay schedules but one.
For this problem, a properly tuned exponential decay seems like a very efficient solution,
and our dynamic tuning captures this solution. Arguably, five runs with different exponential decays might outperform our dynamic approach, but this would critically depend on the chosen bounds for the parameters and luck in the design of experiments. Standard BO (over the parameters) might be an alternative, but five observations would be too small to run it.

\section{Multi-Task Scheduling}\label{sec:multitask}
\subsection{Multi-Task Learning}
We now consider the case where $\Omega$ is discrete and relatively small (say, $|\Omega| =M \leq 10$), but can be increased when a new task needs to be solved, 
similarly to \cite{poloczek2016warm}. The objective is then to find an optimal set of learning rates rather than a single one. However, as an efficient learning rate schedule for a task is often found to perform well for another, we assume that the values of the set share some resemblance.

Several sampling strategies have been proposed recently in this context  \citep{swersky2013multi,ginsbourger2014bayesian,poloczek2016warm,pearce2018continuous}. 
However, all exploit the fact that posterior distributions are available in closed form. As our model is sampling-based, using those approaches would be either impractical, or overly expensive computationally. Hence, we propose a new principled strategy, 
adapted from the \truvar \ algorithm of \cite{bogunovic2016truncated},
originally proposed for optimisation and level-set estimation. 

We first extend our model to multiple tasks, by indexing the latent GP 
$f$ on $\omega$ on top of $Y$ and $x$. Then, 
following \cite{swersky2013multi}, we assume a product kernel for $f$:
\begin{equation*}
 k_f\left[ (Y, x, \omega), (Y', x', \omega') \right] = k_Y(Y, Y') k_x(x, x') k_\Omega(\omega, \omega').
\end{equation*}
To facilitate inference, we assume further that the tasks can be embedded in a low-dimensional latent space, $\omega \rightarrow w \in \Rset^L$.
This results in a set of $L \times M$ parameters to infer (the locations of the tasks in the latent space), independently of the number of runs.

\subsection{Sequential Infill Strategy}
In a nutshell, \truvar \ repeatedly applies two steps: 1) select a set of reference points $\mathcal{M} \in \Xset$ (e.g. for optimisation, potential maximisers), then 2) find the observation that greedily shrinks the sum of prediction variances at reference points.

We adapt here this strategy to uncover profile optima, that is:
\begin{equation*}
    \Xset^* =\{\x^{i*} = \argmax_{\x \in \Xset} y\left(\x, T_{d}, \omega_i\right)\}_{i=1}^M.
\end{equation*}
The original algorithm selects as reference points $\mathcal{M}$
all the points for which an upper confidence bound (UCB) of the objective is higher than a threshold.
As we work with continuous design spaces, we decided to simplify this step and 
consider for $\mathcal{M}$ the maximisers $\hat \Xset^* = \{\hat \x^{1*}, \ldots, \hat \x^{M*} \}$ of the UCB of the final trace value for each task, that is:
\begin{equation}\label{eq:ucbargmax}
    \hat \x^{i*} = \argmax_{\x \in \Xset} q_\alpha \left[ y\left(\x, T_{d}, \omega_i\right) \right],  1 \leq i \leq M,  
\end{equation}
with $\alpha \in (0.5, 1)$ so that the quantile defines a UCB for $y(\x, t)$.
Note that to ensure theoretical guarantees, UCB strategies generally require quantile orders that increase with time \citep{kaufmann2012bayesian}.
However, a constant value usually works best in practice \citep{srinivas2010gaussian,bogunovic2018adversarially}, so we focus on this case here.

Due to the lack of data, the performance at $\hat \Xset^*$ is uncertain, which can be quantified by the mean of variances at $\hat \Xset^*$:
\begin{equation*}
 J = \frac{1}{M} \sum_{i=1}^M \text{Var} \left[ y \left(\hat \x^{i*}, T_d, \omega_i \right) \right].
\end{equation*}
Note that as $\hat \Xset^*$ is chosen using a UCB, it is likely to correspond to values for which the model has a high prediction variance. So, $J$ may increase monotonically with $\alpha$, which acts as a tuning parameter for the exploration / exploitation trade-off.

Now, we would like to find the run (learning rate and task) that reduces $J$ the most. Assume a potential candidate $(\x, \omega)$, that would provide, if evaluated, an additional set of observations $\y_{\x, \omega}$.
Conditioning the model on this new data would reduce the prediction uncertainty at $\hat \Xset^*$ (by law of total variance), which we can measure with 
\begin{eqnarray*}
  \bar J(\x, \omega) = \frac{1}{M} \sum_{i=1}^M \text{Var} \left[ y\left(\hat \x^{i*}, T_d, \omega_i \right) | \y_{\x, \omega} \right] \leq J,
\end{eqnarray*}
where $\text{Var} \left(. | \y_{\x, \omega} \right)$ denotes the variance conditionally on $\y_{\x, \omega}$.
In the case of regular GP models, $\bar J$ is actually available in closed form 
independently of the values of $\y_{\x, \omega}$ \citep{bogunovic2016truncated}.
This is not the case here,
so we replace $\bar J(\x, \omega)$ by its
expectation over the values of $\y_{\x, \omega}$, which leads to the following sampling strategy:
\begin{equation}\label{eq:suropt}
    \{ \x\new, \omega\new \}= \argmin_{\{\x, \omega \}} \esp_{\y_{\x, \omega}}  \left( \bar J(\x, \omega) \right).
\end{equation}

In practice, this criterion is not available in closed form, and must be computed using a double Monte-Carlo loop. 
However, conditioning on $\y_{\x, \omega}$ can be approximated simply, as follow.
First, samples of $\{f(Y_0^\text{new}, x_1^\text{new}), \ldots, f(Y_{d-1}^\text{new}, x_d^\text{new})\}$ are obtained recursively, as in \cref{sec:prediction}.
Then, conditioning $f$ on each of those samples and computing the new conditional variance as in \cref{sec:prediction} allows to compute \cref{eq:suropt}.

Once $\x\new$ and $\omega\new$ are obtained, the corresponding experiment is run and the model is updated, which in turn leads to a new set $\hat \Xset^*$, etc. Once the budget is exhausted, the final set $\hat \Xset^*$ may be chosen using a different $\alpha$ (either 0.5 for a risk-neutral solution or $\leq 0.5$ for a risk-averse one). The pseudo-code of the strategy is given in alg. \ref{alg:multitask}.

\begin{algorithm}[H]
    \caption{Multi-task tuning}\label{alg:multitask}
    Choose $N_0$, $N$, $d$, $T_d$, $L$, $\alpha$\;
    Select initial set of experiments $\{\x_0, \omega_0\}, \ldots, \{\x_{N_0}, \omega_{N_0}\}$\;
    Run $N_0$ SGDs for $T_d$ steps\;
    Gather $N_0$ traces, build trace model\;
    \For{$i \gets N_0+1$ to $N$}{
      Find the optimistic set $\hat \Xset^*$ by solving \cref{eq:ucbargmax}\;
      Find the experiment $\{\x\new, \omega\new \}$ that reduces the uncertainty related to $\hat \Xset^*$ by solving \cref{eq:suropt}\;
      Run SGD with $\{\x\new, \omega\new \}$\;
      Gather new trace, update trace model\;
    }
    Return the set of optimal learning rates $\hat \Xset^*$\;
\end{algorithm}

Following \cite{wilson2018maximizing}, we use the so-called \textit{reparametrisation trick} when generating samples in order to solve \cref{eq:suropt} with respect to $\x_{new}$ using gradient-based optimisation. Note that $\omega_{new}$ is found by exhaustive search. 

\subsection{Experiment: Multi-Task Setting with SVGP on MNIST}
\label{sec:exp:mnist}
To illustrate our strategy, we consider the following setup. 
We use the MNIST digit dataset, split into five binary classification problems (0 against 1, 2 against 3 and so on until 8 against 9). Our goal is to fit a sparse GP classification model to each of these datasets, by maximising its ELBO using the Adam algorithm.

We choose here $T_d = 1,000$ Adam iterations, and $d=5$ different learning rate values. The learning rates are bounded by $[10^{-5}, 10^{-3}]$. 
$N_0=5$ initial runs are performed with learning rates chosen by Latin Hypercube sampling (in the logarithmic space), and $15$ runs are added sequentially according to our \truvar \ strategy.
The GP model for $f$ uses a Mat\'ern-$5/2$ kernel, an exponential link function, $50$ inducing points and a dimension $L=2$ for the latent space of tasks.
To ease the resolution of \cref{eq:ucbargmax}, we follow a greedy approach by searching for one learning rate value at a time, starting from $x_0$, which is in line with the Markov assumption of the model.

\begin{figure}[htbp]
 \begin{minipage}[c]{\textwidth}
     \includegraphics[trim= 0mm 124mm 0mm 8mm, clip, width=.8\textwidth]{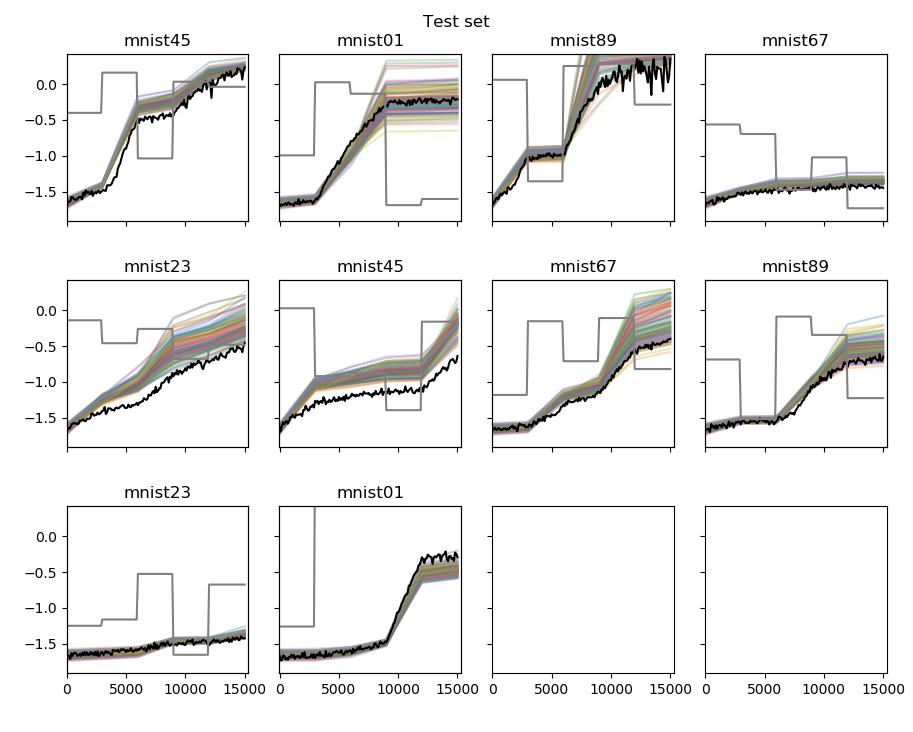}
     \includegraphics[trim= 15mm 65mm 165mm 60mm, clip, width=.175\textwidth]{fig/test_multi.png}\\
     \begin{minipage}[c]{.61\textwidth}
         \includegraphics[trim= 0mm 64mm 0mm 2mm, clip, width=\textwidth]{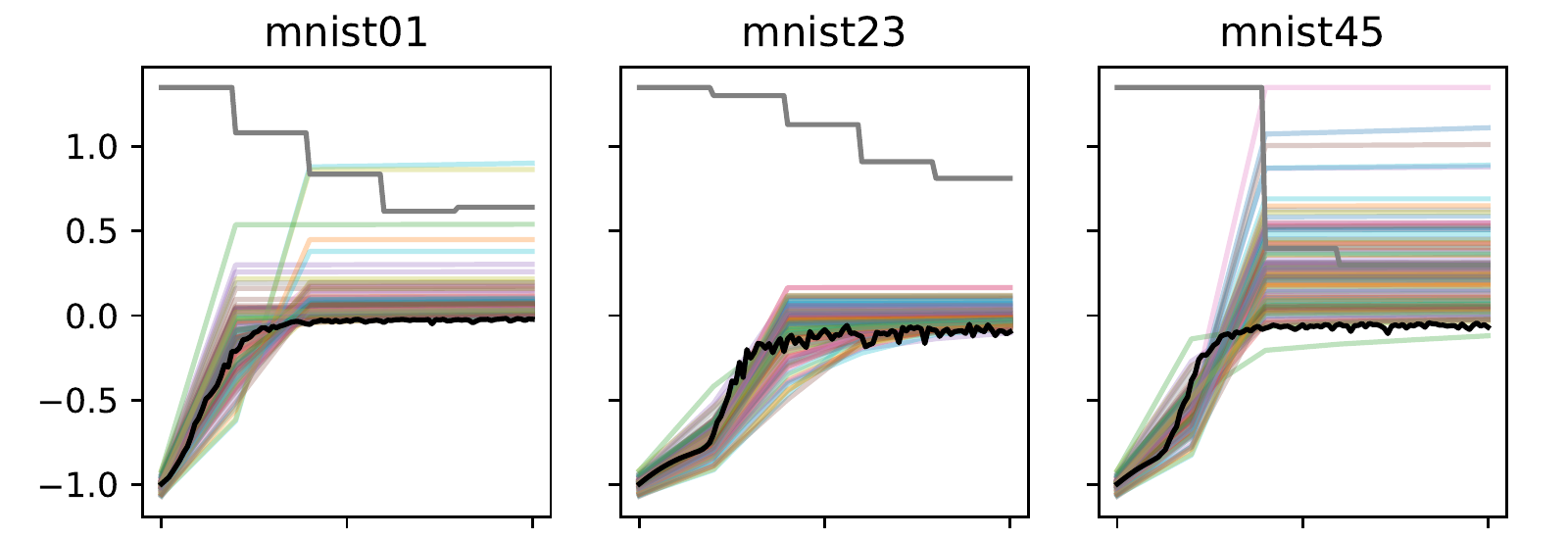}\\
         \includegraphics[trim= 0mm 0mm 0mm 114mm, clip, width=\textwidth]{fig/binary_minst_multitask.pdf}
     \end{minipage} \hspace{-1mm}
     \begin{minipage}[c]{0.37\textwidth}
         \includegraphics[trim= 14mm 0mm 50mm 60mm, clip, width=\textwidth]{fig/binary_minst_multitask.pdf}
     \end{minipage}
 \end{minipage}
 \caption{Actual traces (black), predictions (color) and learning rates (grey, in log-scale between $10^{-5}$ and $10^{-3}$). The learning rates are randomly chosen for the top row, and set to their optimal estimates on the bottom right.} \label{fig:multiresults}
\end{figure}

\Cref{fig:multiresults} shows the learning rate profiles and corresponding runs obtained after running the procedure, as well as some predictions for randomly chosen learning rates. We first observe the flexibility of our model, which is able to capture complex traces while providing relevant uncertainty estimates (top plots).
Then, for all tasks, the learning rates found reach the upper bound at first and decreases when the trace reaches a plateau. The optimal way of decreasing the learning rate depends on the task. One can see that the predictions are uncertain, but only on one side (some samples largely overestimate the true trace but median ones are quite close to the truth).

\begin{figure}[htbp]
 \centering
    \begin{minipage}[c]{.5\linewidth}
    \includegraphics[trim= 0mm 3mm 0mm 5mm, clip, width=.95\textwidth]{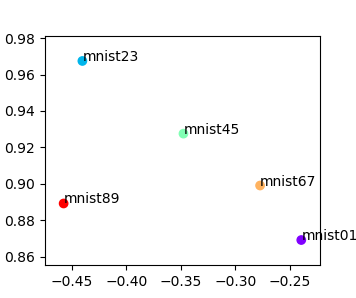}
    \end{minipage}
 \caption{Latent variable values for each dataset.} \label{fig:multiresultslatent}
\end{figure}

\Cref{fig:multiresultslatent} shows the estimated proximity between tasks. 
Here, all the tasks have been found relatively similar, as they all lead to 
close learning rate schedules. 
One may notice for instance that mnist01 is at an edge of the domain, 
which can be imputed to a different initial trace behaviour.

\subsection{Extension: Warm-Starting SGD for a New Task}\label{sec:newtask}
Now, assume that a new task $\omega_{new}$ is added to the current set $\Omega$.
Unfortunately, our model cannot be used directly as the value of the corresponding latent variables $w_{new}$ is unknown.
The first solution is to find a ``universal'' tuning: this can be obtained by 
maximising the prediction of $y\left(\x, T_d, \omega_{new}\right)$ averaged over all possible values for $w_{new}$. 
This average can be calculated by Monte-Carlo assuming a probability measure for $w_{new}$, 
for instance the Lebesgue measure over the convex hull of $w_1, \ldots, w_M$. 

Alternatively, one might want to spend some computing budget (say, $T_i$, $i>0$) to learn $w_{new}$ and achieve then a better learning rate tuning. Assuming again a measure for $w_{new}$, 
an informative experiment would correspond to a learning rate for which the proportion of the variance of the predictor due to the uncertainty on $w_{new}$ is maximal. Averaging over all time steps, we define our sampling strategy (with again a criterion computable by Monte-Carlo) as:
\begin{equation*}
 \x_{:i}^* = \argmax_{\x_{:i} \in [0,1]^i} \sum_{t=0}^{T_i}{\frac{\esp_w \left[ \var(Y(\x_{:i}, t, w) | w) \right]} {\var(Y(\x_{:i}, t, w)}},
\end{equation*}
with $\x_{:i} = [x_0, \ldots, x_i]$.
Note that once $w_{new}$ is estimated, it is possible to apply alg. \ref{alg:coldstart}, 
exploiting the flexibility of dynamic tuning while leveraging information from previous runs.
A more integrated approach would use directly alg. \ref{alg:coldstart} while measuring and accounting for uncertainty in $w_{new}$; this is left for future work.

\section{Concluding Comments}
We proposed a probabilistic model for the traces of optimisers, which input parameters (the choice of learning rate values)
correspond to particular periods of the optimisation. This allowed us to define a versatile framework to tackle a set of problems:
tuning the optimiser for a set of similar tasks, warm-starting it for a new task or on-line adaptation of the learning rate for a
cold-started run.

Convergence proof for the multitask strategy has not been considered here. 
We believe that the results of \cite{bogunovic2016truncated} may be adapted to our case: 
this is left for future work. 
Other possible extensions are to apply our framework to other optimisers: for instance, 
to control the population sizes of evolutionary strategy algorithms such as CMAES \citep{hansen2001completely}, for which adaptation mechanisms have been found promising \citep{nishida2018psacmaes}. 
Finally, additional efficiency could be achieved by leveraging the use of varying dataset size, in the spirit of \cite{klein2017fast,falkner2018bohb} for instance.


\begin{thebibliography}{42}
\providecommand{\natexlab}[1]{#1}
\providecommand{\url}[1]{\texttt{#1}}
\providecommand{\urlprefix}{}

\bibitem[{Andrychowicz et~al.(2016)Andrychowicz, Denil, Gomez, Hoffman, Pfau,
  Schaul, Shillingford, and De~Freitas}]{andrychowicz2016learning}
Andrychowicz, M., Denil, M., Gomez, S., Hoffman, M.W., Pfau, D., Schaul, T.,
  Shillingford, B., De~Freitas, N.: Learning to learn by gradient descent by
  gradient descent.
\newblock In: Advances in Neural Information Processing Systems. pp. 3981--3989
  (2016)

\bibitem[{Baydin et~al.(2017)Baydin, Cornish, Rubio, Schmidt, and
  Wood}]{baydin2017online}
Baydin, A.G., Cornish, R., Rubio, D.M., Schmidt, M., Wood, F.: Online learning
  rate adaptation with hypergradient descent.
\newblock arXiv preprint arXiv:1703.04782  (2017)

\bibitem[{Bengio(2012)}]{bengio2012practical}
Bengio, Y.: Practical recommendations for gradient-based training of deep
  architectures.
\newblock In: Neural networks: Tricks of the trade, pp. 437--478. Springer
  (2012)

\bibitem[{Bergstra et~al.(2011)Bergstra, Bardenet, Bengio, and
  K{\'e}gl}]{bergstra2011algorithms}
Bergstra, J.S., Bardenet, R., Bengio, Y., K{\'e}gl, B.: Algorithms for
  hyper-parameter optimization.
\newblock In: Advances in neural information processing systems. pp. 2546--2554
  (2011)

\bibitem[{Bogunovic et~al.(2018)Bogunovic, Scarlett, Jegelka, and
  Cevher}]{bogunovic2018adversarially}
Bogunovic, I., Scarlett, J., Jegelka, S., Cevher, V.: Adversarially robust
  optimization with {G}aussian processes.
\newblock In: Advances in Neural Information Processing Systems. pp. 5760--5770
  (2018)

\bibitem[{Bogunovic et~al.(2016)Bogunovic, Scarlett, Krause, and
  Cevher}]{bogunovic2016truncated}
Bogunovic, I., Scarlett, J., Krause, A., Cevher, V.: Truncated variance
  reduction: A unified approach to {B}ayesian optimization and level-set
  estimation.
\newblock In: Advances in neural information processing systems. pp. 1507--1515
  (2016)

\bibitem[{Chollet(2009)}]{resnetkeras}
Chollet, F.: Keras implementation of {R}es{N}et for {CIFAR}.
\newblock \url{https://keras.io/examples/cifar10_resnet/} (2009)

\bibitem[{Domhan et~al.(2015)Domhan, Springenberg, and
  Hutter}]{domhan2015speeding}
Domhan, T., Springenberg, J.T., Hutter, F.: Speeding up automatic
  hyperparameter optimization of deep neural networks by extrapolation of
  learning curves.
\newblock In: Twenty-Fourth International Joint Conference on Artificial
  Intelligence (2015)

\bibitem[{Duchi et~al.(2011)Duchi, Hazan, and Singer}]{duchi2011adaptive}
Duchi, J., Hazan, E., Singer, Y.: Adaptive subgradient methods for online
  learning and stochastic optimization.
\newblock Journal of Machine Learning Research 12(Jul), 2121--2159 (2011)

\bibitem[{Falkner et~al.(2018)Falkner, Klein, and Hutter}]{falkner2018bohb}
Falkner, S., Klein, A., Hutter, F.: Bohb: Robust and efficient hyperparameter
  optimization at scale.
\newblock arXiv preprint arXiv:1807.01774  (2018)

\bibitem[{Ginsbourger et~al.(2014)Ginsbourger, Baccou, Chevalier, Perales,
  Garland, and Monerie}]{ginsbourger2014bayesian}
Ginsbourger, D., Baccou, J., Chevalier, C., Perales, F., Garland, N., Monerie,
  Y.: {B}ayesian adaptive reconstruction of profile optima and optimizers.
\newblock SIAM/ASA Journal on Uncertainty Quantification 2(1), 490--510 (2014)

\bibitem[{Gugger and Howard(2018)}]{gugger2018}
Gugger, S., Howard, J.: Adamw and super-convergence is now the fastest way to
  train neural nets (Jul 2018),
  \urlprefix\url{https://www.fast.ai/2018/07/02/adam-weight-decay/}

\bibitem[{Hansen and Ostermeier(2001)}]{hansen2001completely}
Hansen, N., Ostermeier, A.: Completely derandomized self-adaptation in
  evolution strategies.
\newblock Evolutionary computation 9(2), 159--195 (2001)

\bibitem[{He et~al.(2016)He, Zhang, Ren, and Sun}]{he2016deep}
He, K., Zhang, X., Ren, S., Sun, J.: Deep residual learning for image
  recognition.
\newblock In: Proceedings of the IEEE conference on computer vision and pattern
  recognition. pp. 770--778 (2016)

\bibitem[{Hensman et~al.(2015)Hensman, Matthews, and
  Ghahramani}]{hensman2015scalable}
Hensman, J., Matthews, A.G.d.G., Ghahramani, Z.: Scalable variational
  {G}aussian process classification.
\newblock In: Proceedings of the Eighteenth International Conference on
  Artificial Intelligence and Statistics (2015)

\bibitem[{Hoffman et~al.(2013)Hoffman, Blei, Wang, and Paisley}]{hoffman2013}
Hoffman, M.D., Blei, D.M., Wang, C., Paisley, J.: {Stochastic Variational
  Inference}.
\newblock Journal of Machine Learning Research  (2013)

\bibitem[{Kaufmann et~al.(2012)Kaufmann, Capp{\'e}, and
  Garivier}]{kaufmann2012bayesian}
Kaufmann, E., Capp{\'e}, O., Garivier, A.: On {B}ayesian upper confidence
  bounds for bandit problems.
\newblock In: Artificial intelligence and statistics. pp. 592--600 (2012)

\bibitem[{Klein et~al.(2017{\natexlab{a}})Klein, Falkner, Bartels, Hennig, and
  Hutter}]{klein2017fast}
Klein, A., Falkner, S., Bartels, S., Hennig, P., Hutter, F.: Fast {B}ayesian
  optimization of machine learning hyperparameters on large datasets.
\newblock In: International Conference on Artificial Intelligence and
  Statistics (AISTATS 2017). pp. 528--536. PMLR (2017{\natexlab{a}})

\bibitem[{Klein et~al.(2017{\natexlab{b}})Klein, Falkner, Springenberg, and
  Hutter}]{klein2016learning}
Klein, A., Falkner, S., Springenberg, J.T., Hutter, F.: Learning curve
  prediction with {B}ayesian neural networks.
\newblock In: ICLR (2017{\natexlab{b}})

\bibitem[{Krizhevsky and Hinton(2009)}]{krizhevsky2009learning}
Krizhevsky, A., Hinton, G.: Learning multiple layers of features from tiny
  images.
\newblock Tech. rep., Citeseer (2009)

\bibitem[{Leontaritis and Billings(1985)}]{leontaritis1985input}
Leontaritis, I., Billings, S.A.: Input-output parametric models for non-linear
  systems part ii: stochastic non-linear systems.
\newblock International journal of control 41(2), 329--344 (1985)

\bibitem[{Li et~al.(2018)Li, Jamieson, DeSalvo, Rostamizadeh, and
  Talwalkar}]{li2018hyperband}
Li, L., Jamieson, K., DeSalvo, G., Rostamizadeh, A., Talwalkar, A.: Hyperband:
  A novel bandit-based approach to hyperparameter optimization.
\newblock Journal of Machine Learning Research 18(185), 1--52 (2018)

\bibitem[{Loshchilov and Hutter(2019)}]{loshchilov2018decoupled}
Loshchilov, I., Hutter, F.: Decoupled weight decay regularization.
\newblock In: ICLR (2019)

\bibitem[{Matthews et~al.(2016)Matthews, Hensman, Turner, and
  Ghahramani}]{matthews2016sparse}
Matthews, A.G.d.G., Hensman, J., Turner, R., Ghahramani, Z.: On sparse
  variational methods and the kullback-leibler divergence between stochastic
  {P}rocesses.
\newblock Journal of Machine Learning Research 51, 231--239 (2016)

\bibitem[{Matthews et~al.(2017)Matthews, Van Der~Wilk, Nickson, Fujii,
  Boukouvalas, Le{\'o}n-Villagr{\'a}, Ghahramani, and
  Hensman}]{matthews2017gpflow}
Matthews, A.G.d.G., Van Der~Wilk, M., Nickson, T., Fujii, K., Boukouvalas, A.,
  Le{\'o}n-Villagr{\'a}, P., Ghahramani, Z., Hensman, J.: Gpflow: A {G}aussian
  {P}rocess library using tensorflow.
\newblock The Journal of Machine Learning Research 18(1), 1299--1304 (2017)

\bibitem[{Nishida and Akimoto(2018)Nishida, K., and Akimoto, Y.}]{nishida2018psacmaes}
Nishida, K., Akimoto, Y.: PSA-CMA-ES: CMA-ES with population size adaptation. 
\newblock In: Proceedings of the Genetic and Evolutionary Computation Conference (pp. 865-872) (2018)

\bibitem[{Pearce and Branke(2018)}]{pearce2018continuous}
Pearce, M., Branke, J.: Continuous multi-task {B}ayesian optimisation with
  correlation.
\newblock European Journal of Operational Research 270(3), 1074--1085 (2018)

\bibitem[{Picheny and Ginsbourger(2013)}]{picheny2013nonstationary}
Picheny, V., Ginsbourger, D.: A nonstationary space-time {G}aussian {P}rocess
  model for partially converged simulations.
\newblock SIAM/ASA Journal on Uncertainty Quantification 1(1), 57--78 (2013)

\bibitem[{Poloczek et~al.(2016)Poloczek, Wang, and Frazier}]{poloczek2016warm}
Poloczek, M., Wang, J., Frazier, P.I.: Warm starting {B}ayesian optimization.
\newblock In: Proceedings of the 2016 Winter Simulation Conference. pp.
  770--781. IEEE Press (2016)

\bibitem[{Reddi et~al.(2018)Reddi, Kale, and Kumar}]{reddi2018convergence}
Reddi, S.J., Kale, S., Kumar, S.: On the convergence of {ADAM} and beyond.
\newblock In: ICLR (2018)

\bibitem[{Saul et~al.(2016)Saul, Hensman, Vehtari, Lawrence
  et~al.}]{saul2016chained}
Saul, A.D., Hensman, J., Vehtari, A., Lawrence, N.D., et~al.: Chained
  {G}aussian {P}rocesses.
\newblock In: AISTATS. pp. 1431--1440 (2016)

\bibitem[{Shahriari et~al.(2016)Shahriari, Swersky, Wang, Adams, and
  De~Freitas}]{shahriari2016taking}
Shahriari, B., Swersky, K., Wang, Z., Adams, R.P., De~Freitas, N.: Taking the
  human out of the loop: A review of {B}ayesian optimization.
\newblock Proceedings of the IEEE 104(1), 148--175 (2016)

\bibitem[{Smith(2017)}]{smith2017cyclical}
Smith, L.N.: Cyclical learning rates for training neural networks.
\newblock In: 2017 IEEE Winter Conference on Applications of Computer Vision
  (WACV). pp. 464--472. IEEE (2017)

\bibitem[{Smith and Topin(2019)}]{smith2019super}
Smith, L.N., Topin, N.: Super-convergence: Very fast training of neural
  networks using large learning rates.
\newblock In: Artificial Intelligence and Machine Learning for Multi-Domain
  Operations Applications. vol. 11006, p. 1100612. International Society for
  Optics and Photonics (2019)

\bibitem[{Snoek et~al.(2012)Snoek, Larochelle, and Adams}]{snoek2012practical}
Snoek, J., Larochelle, H., Adams, R.P.: Practical {B}ayesian optimization of
  machine learning algorithms.
\newblock In: Advances in neural information processing systems. pp. 2951--2959
  (2012)

\bibitem[{Srinivas et~al.(2010)Srinivas, Krause, Kakade, and
  Seeger}]{srinivas2010gaussian}
Srinivas, N., Krause, A., Kakade, S., Seeger, M.: {G}aussian {P}rocess
  optimization in the bandit setting: no regret and experimental design.
\newblock In: Proceedings of the 27th International Conference on International
  Conference on Machine Learning. pp. 1015--1022. Omnipress (2010)

\bibitem[{Swersky et~al.(2013)Swersky, Snoek, and Adams}]{swersky2013multi}
Swersky, K., Snoek, J., Adams, R.P.: Multi-task {B}ayesian optimization.
\newblock In: Advances in neural information processing systems. pp. 2004--2012
  (2013)

\bibitem[{Swersky et~al.(2014)Swersky, Snoek, and Adams}]{swersky2014freeze}
Swersky, K., Snoek, J., Adams, R.P.: Freeze-thaw {B}ayesian optimization.
\newblock arXiv preprint arXiv:1406.3896  (2014)

\bibitem[{Tieleman and Hinton(2012)}]{tieleman2012lecture}
Tieleman, T., Hinton, G.: Lecture 6.5-rmsprop: Divide the gradient by a running
  average of its recent magnitude.
\newblock COURSERA: Neural networks for machine learning 4(2), 26--31 (2012)

\bibitem[{Titsias(2009)}]{titsias2009}
Titsias, M.: {Variational Learning of Inducing Variables in Sparse {G}aussian
  Processes}.
\newblock Artificial Intelligence and Statistics  (2009)

\bibitem[{{van der Wilk} et~al.(2020){van der Wilk}, Dutordoir, John, Artemev,
  Adam, and Hensman}]{GPflow2020multioutput}
{van der Wilk}, M., Dutordoir, V., John, S., Artemev, A., Adam, V., Hensman,
  J.: A framework for interdomain and multioutput {G}aussian processes.
\newblock arXiv:2003.01115  (2020),
  \urlprefix\url{https://arxiv.org/abs/2003.01115}

\bibitem[{Wilson et~al.(2018)Wilson, Hutter, and
  Deisenroth}]{wilson2018maximizing}
Wilson, J., Hutter, F., Deisenroth, M.: Maximizing acquisition functions for
  {B}ayesian optimization.
\newblock In: Advances in Neural Information Processing Systems. pp. 9884--9895
  (2018)

\end{thebibliography}
\end{document}